\documentclass[letterpaper, 10 pt, conference]{ieeeconf}  % Comment this line out if you need a4paper

\IEEEoverridecommandlockouts                              % This command is only needed if 
\usepackage{graphics,graphicx} % for pdf, bitmapped graphics files
\usepackage{bm, amsmath} % assumes amsmath package installed
\usepackage{cite}
\usepackage{array}
\usepackage{paralist}
\usepackage{multirow}
\usepackage[table]{xcolor}% http://ctan.org/pkg/xcolor

\usepackage{caption}
\usepackage{subcaption}
\usepackage{tikz}
\usepgflibrary{arrows}

\title{\LARGE \bf
Design Paradigms Based on Spring Agonists for \\ Underactuated Robot Hands: Concepts and Application$^{*}$
}

\author{Tianjian Chen$^{\dagger}$, Tianyi Zhang$^{\dagger}$, and  Matei Ciocarlie% <-this % stops a space
\thanks{$^*$This work was supported by the NASA Early Stage Innovations (ESI) Program through award NNX16AD13G.}% <-this % stops a space
\thanks{$^\dagger$Authors have contributed equally to this work.}
\thanks{All authors are with the department of Mechanical Engineering, Columbia University, New York, NY 10027, USA.
        {\{tianjian.chen, tianyi.zhang2, matei.ciocarlie\}@columbia.edu}}%
}

\begin{document}

\setlength{\abovedisplayskip}{2mm} 
\setlength{\belowdisplayskip}{2mm} 

\maketitle
\thispagestyle{empty}
\pagestyle{empty}

%%%%%%%%%%%%%%%%%%%%%%%%%%%%%%%%%%%%%%%%%%%%%%%%%%%%%%%%%%%%%%%%%%%%%%%%%%%%%%%%
\begin{abstract}
In this paper, we focus on a rarely used paradigm in the design of underactuated robot hands: the use of springs as agonists and tendons as antagonists. We formalize this approach in a design matrix also considering its interplay with the underactuation method used (one tendon for multiple joints vs. multiple tendons on one motor shaft). We then show how different cells in this design matrix can be combined in order to facilitate the implementation of desired postural synergies with a single motor. Furthermore, we show that when agonist and antagonist tendons are combined on the same motor shaft, the resulting spring force cancellation can be leveraged to produce multiple desirable behaviors, which we demonstrate in a physical prototype.
\end{abstract}

\section{Introduction}
\label{sec:intro}

Inspired by biological musculoskeletal systems, tendon-driven mechanisms have numerous benefits in robotics: they allow actuators to be located close to the base, while remaining compact, lightweight, and flexible in installation \cite{tsai1995design}. Due to these advantages, tendon-driven robotic hands have drawn significant attention \cite{jacobsen1986design, loucks1987modeling, lovchik1999robonaut, shadowhand, deshpande2011mechanisms, xu2016design, gosselin2008anthropomorphic, dollar2010highly, wang2011highly, odhner2014compliant, ciocarlie2014velo, catalano2014adaptive, stuart2017ocean}. Of particular interest is the category of \textit{underactuated} tendon-driven hands \cite{gosselin2008anthropomorphic, dollar2010highly, wang2011highly, odhner2014compliant, ciocarlie2014velo, catalano2014adaptive, stuart2017ocean}. By running a tendon through multiple joints with positive moment arms, it is straightforward to couple multiple joints and actuate them to grasp in a synergistic fashion. Furthermore, multiple tendons can be connected to a single motor shaft, allowing for an additional type of joint coupling.

However, for all the advantages above, tendon-driven mechanisms are not without shortcomings. Of particular concern to this study, tendons can only pull. To enable bidirectional motion, the tendon can be replaced by a belt, or two motors can be used, one for each direction; however, both options negate the compactness that is so attractive in the use of tendons in the first place. 

In robot hands, particularly underactuated ones, designers often choose a different option: use a passive spring at each joint to account for one direction of motion. This means that only one direction is actively actuated, while the other is passive. However, that is generally acceptable for hands where movement in one direction (closing to grasp) is of more interest than its opposite (opening to release). Using the terminology of human physiology, we define the mechanism generating movement in the direction of interest as the ``\textit{agonist}'', and the opposite side as the ``\textit{antagonist}''. In the rest of the study we will consider finger flexion and adduction to represent the directions of interest for grasping.

\begin{figure}[t]
%    \footnotesize
%    \begin{tabular}{>{\centering\arraybackslash}m{0.12\columnwidth}	>{\centering\arraybackslash}m{0.35\columnwidth} >{\centering\arraybackslash}m{0.35\columnwidth} }
%      &  \textit{Tendons as Agonists (\textbf{TA})} &  \textit{Springs as Agonists (\textbf{SA})}  \\ 
%      \textit{One Tendon across Multiple Joints (\textbf{OTMJ})} &  \vspace{2mm}   \hspace{5mm}\includegraphics[width=0.5\linewidth]{images/cover1.pdf} \includegraphics[width=0.15\linewidth]{images/check_mark.png} & \vspace{2mm} \includegraphics[width=0.5\linewidth]{images/cover2.pdf} \\  
%      \textit{One Tendon for One Joint (\textbf{OTOJ})} & \vspace{2mm} \includegraphics[width=0.5\linewidth]{images/cover3.pdf} & \vspace{2mm} \hspace{5mm} \includegraphics[width=0.5\linewidth]{images/cover4.pdf} \includegraphics[width=0.15\linewidth]{images/check_mark.png} \\ 
%    \end{tabular}
    % \centering    
    % \includegraphics[width=0.7\linewidth]{images/cover_legend.pdf}
%    \centering  
%    \begin{tikzpicture}[thick]
%        \draw [black, line width=2mm, preaction={-triangle 60,thick,draw, line width=1mm, shorten >=-3mm}] (0, 4mm) -- (0, 0) node [right] {};
%    \end{tikzpicture}
    % \vspace{3mm}
   
    \begin{tabular}{cc}
         \includegraphics[width=0.5\linewidth]{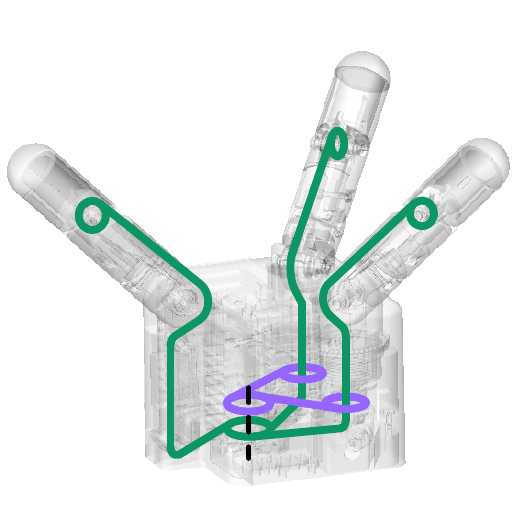} & \includegraphics[width=0.5\linewidth]{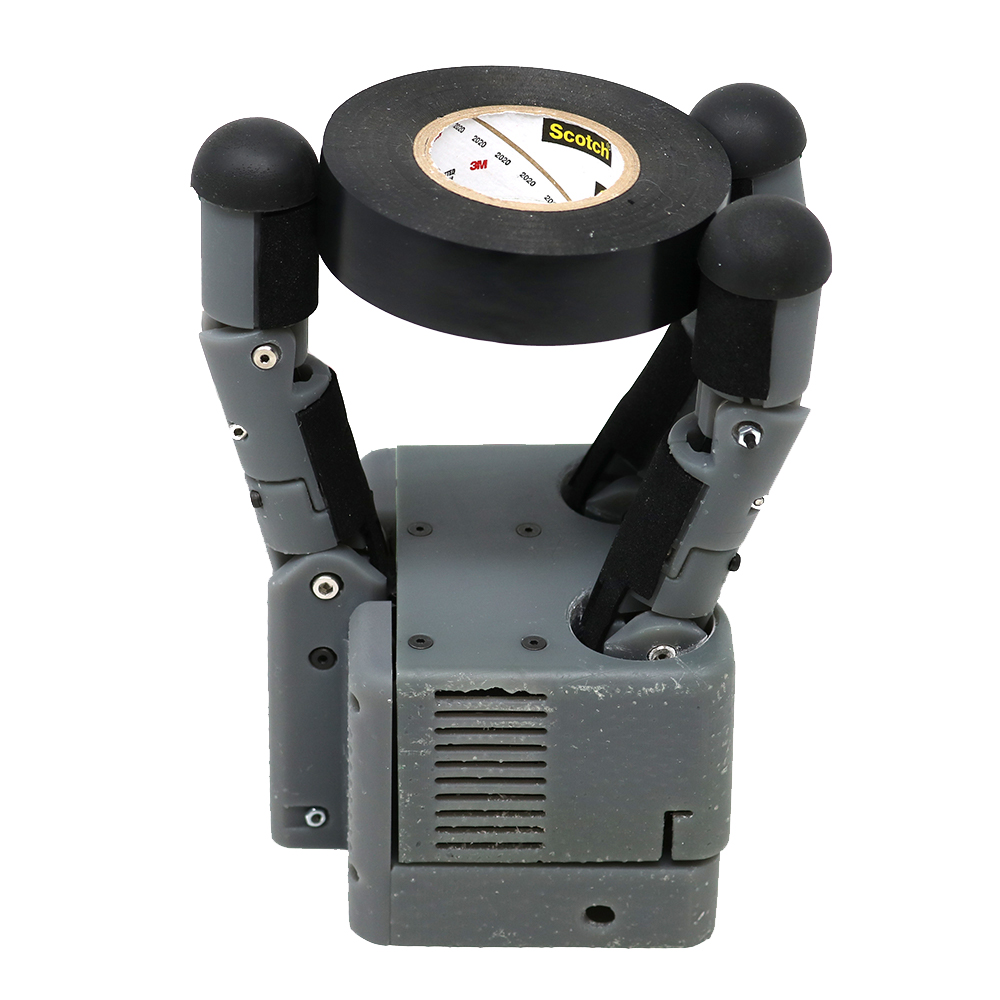} \\
    \end{tabular}
    \normalsize
    \caption{A robot hand using both ``spring agonists + tendon antagonists'' (for abduction-adduction) and ``tendon agonists + spring antagonists'' (for flexion-extension). Tendons acting in both roles are combined on the same motor shaft. As a result, the design can be optimized to achieve desired postural synergies with a single motor, but also to exhibit other desirable behaviors: the hand can maintain a grasp pose even without power, but is still backdrivable and can be shaped by human intervention if needed.}
    \label{fig:cover}
    \vspace{-5mm}

\end{figure}

\begin{table*}[t!]\centering
\caption{Design matrix of the tendon transmission in underactuated hands}

\label{tab:design_matrix}
    \begin{tabular}{| >{\centering\arraybackslash}m{0.07\linewidth} | >{\arraybackslash}p{0.22\linewidth}  >{\arraybackslash}p{0.18\linewidth} | >{\arraybackslash}p{0.22\linewidth}  >{\arraybackslash}p{0.18\linewidth} |}
    \hline
        & \multicolumn{2}{c|}{\textbf{Tendons as Agonists (TA)}}  & \multicolumn{2}{c|}{\textbf{Springs as Agonists (SA)}} \\ 
    \hline
    \vspace{5mm}
    \multirow{10}{\linewidth}{\textbf{Multiple Joints per Tendon (MJT)}} 
        & \textbf{TA+MJT} & \multirow{10}{*}{\includegraphics[width=0.95\linewidth]{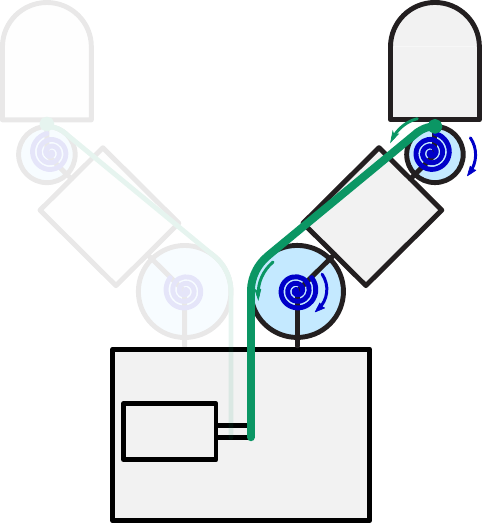}} &\textbf{SA+MJT} & \multirow{10}{*}{\includegraphics[width=0.95\linewidth]{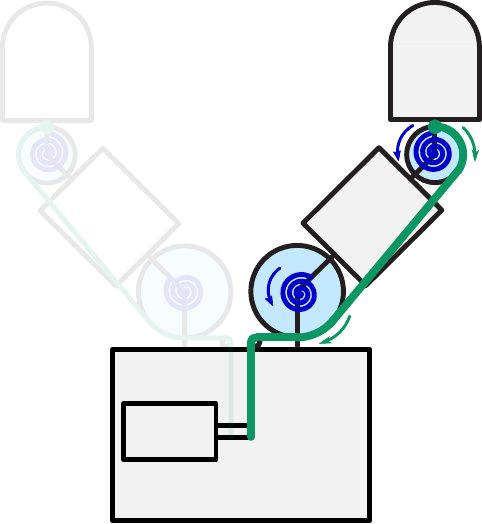}} \\
        & $\bullet$ Soft joint position coupling before contact [due to MJT] & & $\bullet$ Soft joint position coupling before contact [due to MJT] & \\
        & $\bullet$ Joint position differential, breakaway effect (i.e.distal joints can move if proximals are blocked) [due to MJT] & &  $\bullet$ Joint position differential, breakaway effect (i.e.distal joints can move if proximals are blocked) [due to both MJT and SA] & \\
        &  $\bullet$ Joint torque coupling [due to MJT] & &  $\bullet$ Joint torque coupling [due to MJT]  & \\
        & $\bullet$ Grasping force regulation via motor torque [due to TA] & &  $\bullet$ Grasping force regulation is difficult [due to SA] & \\
        & & & & \\ 
    \hline
    \vspace{5mm}
    \multirow{10}{\linewidth}{Single joint per tendon w. \textbf{Multiple Tendons per motor Shaft (MTS)}} 
        &\textbf{TA+MTS} & \multirow{10}{*}{\includegraphics[width=0.95\linewidth]{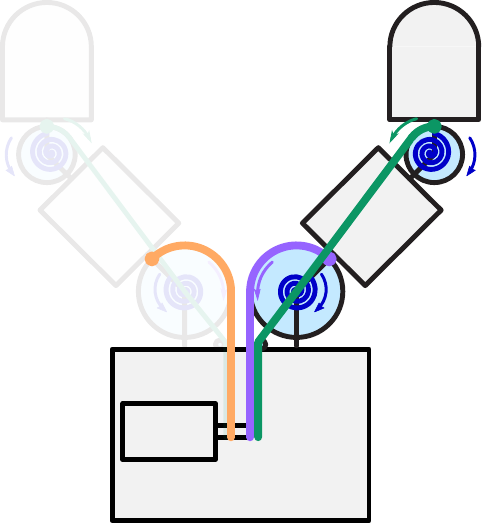}}  & \textbf{SA+MTS} & \multirow{10}{*}{\includegraphics[width=0.95\linewidth]{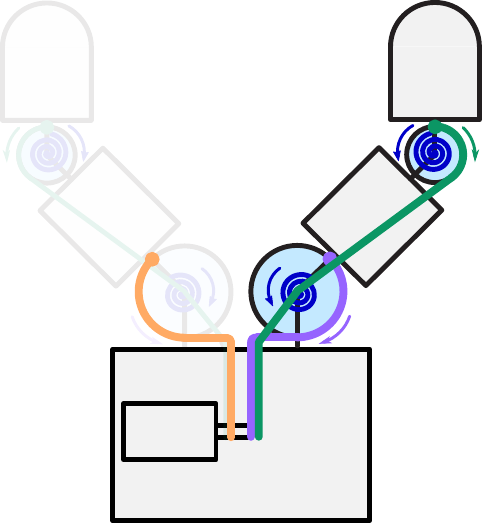}} \\
        & $\bullet$ Precise joint position coupling [due to MTS] & &  $\bullet$ Precise joint position coupling [due to MTS] &  \\
        & $\bullet$ No joint position differential (breakaway) [due to MTS] & &  $\bullet$ Joint position differential, breakaway effect (i.e.some joints can move if others are blocked) [due to SA] & \\
        & $\bullet$ No joint torque coupling, torque differential [due to MTS] & &  $\bullet$ No joint torque coupling, torque differential [due to MTS] & \\
        & $\bullet$ Grasping force regulation via motor torque [due to TA] & &  $\bullet$ Grasping force regulation is difficult [due to SA] &\\
        & & & & \\
        % & & & & \\

\hline
\end{tabular}
%\vspace{2mm}
% (The \textbf{\dag}, \textbf{\ddag}, \textbf{*}, \textbf{\S} symbols before each bullet point show the causal relationship between this point and the corresponding table header.) \\
%(Inside the square brackets after each bullet point we show the cause of that feature.)
%\vspace{-5mm}

\end{table*}

Most existing tendon-driven underactuated hands use active tendons as agonists and passive springs as antagonists~\cite{gosselin2008anthropomorphic, dollar2010highly, wang2011highly, odhner2014compliant, ciocarlie2014velo, catalano2014adaptive, stuart2017ocean}. Furthermore, the most common underactuation scheme is to use a single tendon per motor, routed through multiple joints~\cite{gosselin2008anthropomorphic, dollar2010highly, wang2011highly, odhner2014compliant, ciocarlie2014velo, catalano2014adaptive, stuart2017ocean}. While these paradigms are undoubtedly useful, in this study, we wish to also explore their alternatives: where springs are used as agonists, and where multiple tendons are connected to the same motor shaft. The main contributions of this paper are as follows:
\begin{compactitem}
    \item Combining the two design dimensions mentioned above, we formalize a design matrix for tendon-driven underactuated hands. In one dimension, we have \textit{Tendons as Agonists} (TA) versus \textit{Springs as Agonists} (SA). On the other dimension, we have \textit{Multiple Joints per Tendon} (MJT) versus \textit{single joint per tendon, but Multiple Tendons per motor Shaft} (MTS). This matrix is presented in detail in Table~\ref{tab:design_matrix} and discussed in detail in Sec.~\ref{sec:design_matrix}. We show how different cells in this matrix have different properties, which can be leveraged to obtain the desired behaviors.
    \item Taking ideas from this design matrix, we are the first to present a design combining tendon agonists and spring agonists on different joints in one underactuated hand. Specifically, we propose the use of TA+MJT for flexion/extension and SA+MTS for abduction/adduction. 
    \item We show how our design approach has multiple beneficial effects. First, it facilitates the implementation of grasp synergies in highly underactuated designs. Second, by connecting multiple tendons acting in different roles on the same motor shaft, it allows for optimized spring force cancellation. This in turn allows static equilibrium to be obtained over a wide range of poses. 
    \item We illustrate these concepts on a physical prototype designed for a real use case: manipulation and perching for an Assistive Free Flyer robot on board the International Space Station (ISS). The resulting highly underactuated prototype exhibits versatile grasping behavior via postural synergies, all-pose hand equilibrium without power use, and allows human-intervention when powered off, thanks to spring cancellation effects combined with motor friction.
\end{compactitem}

\section{Related Work}
\label{sec:related_work}

Tendon-driven mechanisms have been commonly used in many different robotic hands for decades, such as the Utah/MIT hand \cite{jacobsen1986design}, the Stanford/JPL hand \cite{loucks1987modeling}, the ACT hand \cite{deshpande2011mechanisms} and the Shadow hand \cite{shadowhand}. These hands provide great dexterity via remote full actuation, but require sophisticated control and are generally considered lack of robustness due to the mechanical complexity. 

Tendon mechanisms turned out to have a bigger role to play when researchers started to look at using them in underactuated  hands. For example, the works from Gosselin et al. \cite{gosselin2008anthropomorphic}, Dollar et al. \cite{dollar2010highly}, Wang et al. \cite{wang2011highly}, Ciocarlie et al. \cite{ciocarlie2014velo}, Odhner et al. \cite{odhner2014compliant}, Catalano et al. \cite{catalano2014adaptive}, Stuart \cite{stuart2017ocean} and Chen et al. \cite{chen2020underactuation} all used tendons to actively drive multiple joints and generate coordination between them. All these examples belong to the category of TA+MJT.

While all the design cases mentioned above have been using tendon to close the hand (TA), springs can also be designed as agonists (SA) in some applications. This idea was used in human-powered hand prosthesis for many years, with the terminology ``voluntary opening'' (reviewed in \cite{smit2012efficiency}). However, in the field of robotic hands, this is rare and under-investigated. The previous work from our group (in collaboration with NASA) \cite{park2017developing} presented a two-finger underactuated tendon-driven gripper for a free-flying robot in the International Space Station (ISS). By using a torsional spring to passively close the hand and an actuated tendon to open the hand, the gripper is able to maintain a grasp when the power is cut off. Also, astronauts can manually open the gripper by countering torques from springs. 
% This example is further discussed in Section \ref{sec:conceptual_design}.

\section{Design Matrix and Actuation Paradigms}
\label{sec:design_matrix}

In this section we present the design matrix in detail, as illustrated in Table~\ref{tab:design_matrix}. We first explain the design choices along each dimension, and then discuss the advantages and disadvantages of each cell in the matrix. We note that the tendons are modeled as inextensible in this paper.

\subsection{Choice of Agonist and Antagonist}

The first dimension of this design matrix is the choice of agonist and antagonist. In the most common way (e.g., \cite{gosselin2008anthropomorphic, dollar2010highly, wang2011highly, odhner2014compliant, ciocarlie2014velo, catalano2014adaptive, stuart2017ocean}), active tendons are routed in the agonistic direction to drive the joint, while springs are used to restore the joint positions (``tendons as agonists (TA)''). The motor collects the tendons to close the fingers by overcoming the spring forces during pre-contact motion, and provides net grasping forces on top of the spring forces after contacts are established. However, this choice can be reversed: the springs can be used as agonists, or the prime mover for grasping, and actively controlled tendons can be used as antagonists (``springs as agonists (SA)''). In this case, the springs are loaded and the tendons are tensioned in one extreme of its range of motion, and the motor releases the tendon to let the springs drive the fingers. 

%The major benefits of this design is that we can \textit{use tendon slack as the simplest breakaway mechanism}: in pre-contact phase, the motor unwinds the tendon to move the finger; whenever a contact is established and a link is blocked, the tendon goes slack and the joint no longer moves together with the motor, which creates a ``breakaway'' (a.k.a. position differential, meaning the transmission can break when needed). 

%However, the disadvantages of this paradigm cannot be neglected: first, the maximum grasping forces are limited by spring forces, which is usually much weaker than the motor forces; second and more importantly, the grasping force regulation is nontrivial --- even though it is possible in theory, it is difficult in practice since it requires accurate motor torque control without changing position (if motor position changes, either the finger gets opened and contacts are lost, or the tendon gets slack and tension drops to zero). 
% Thirdly, in the cases where tendons go slack after grasping, the net grasping torques are determined by position-dependent spring torques, which makes it difficult to design a useful torque relationship for different hand poses.

\subsection{Multi-Joint Coupling}

The second dimension of this design matrix is the coupling scheme across different joints, i.e., the scheme of underactuation. One common paradigm is to have ``Multiple Joints per Tendon (MJT)'' (e.g. \cite{gosselin2008anthropomorphic, dollar2010highly, wang2011highly, odhner2014compliant, ciocarlie2014velo, catalano2014adaptive, stuart2017ocean}). In each joint, the tendon goes around an idler pulley, or simply slides along fixed routing points, creating a torque that can drive the link to move. The alternative is to have a single joint per tendon, but connect "Multiple Tendons per motor Shaft (MTS)", thus allowing a single motor to drive multiple joints.

%This design can achieve strong position coupling that joint positions always follow the a fixed relationship even in the presence of disturbances. Also, the joint torques are not coupled since the forces of different tendons are independent, which can be either an advantage or disadvantage depending on the application.

\subsection{Resulting Design Paradigms}

\subsubsection{TA+MJT} This is the most commonly used paradigm in underactuated hand design, and its advantages are well known. First of all, MJT provides a position differential (breakaway behavior) between joints: whenever a proximal link is blocked by objects, the motor can continue to drive the distal links to close, enabling adaptive grasping. 

It is important to look in more detail at the problem of joint position coupling \textit{before contact}. Based only on the tendon, this coupling is ill-defined: for a given tendon length, and without external forces, there are multiple possible combinations of acceptable joint positions. However, when antagonist springs are added, the problem becomes well-posed: for a given tendon length, the hand will take deterministic joint positions based on spring forces as well as other effects such as joint friction, tendon friction, etc. Thus, there is a "soft" position coupling between joints, one that can be exploited but also one that is sensitive to internal and external disturbances. 

Finally, since the tendon force is shared by multiple joints, the ratio of joint torques from the tendon is always equal to the ratio of tendon moment arms. This feature is a double-edged sword: we can carefully design this ratio and make use of it to make stable grasps, but the constraint that torques must follow this ratio can also be problematic in some applications (we further explain this in our design example in Sec. \ref{sec:conceptual_design}).

\subsubsection{SA+MJT} This paradigm is more rarely used in robotics compared to the one above. Compared to TA+MJT, it has the additional disadvantage that grasping force regulation is difficult: grasping forces are determined by springs when the contacts are made and tendons get slack, and thus can not be controlled by the motor(s). However, it does present the advantage of being able to maintain a grasp in the absence of motor torques~\cite{park2017developing}. Due to the same reason, this paradigm is more well-known and used in the field of human-powered hand prosthetics \cite{smit2012efficiency}.

\subsubsection{TA+MTS} This design is rare. The major limitation is that there is no joint position differential: whenever one joint is blocked by the object, the motor and thus all joints must stop. This results in the incompetence of adapting to object shapes. However, the MTS design still have several notable features: first, it provides precise joint position coupling because the kinematic relationship between joints is well-defined; second, the joint torques are not coupled (i.e., it has torque differential) since joints are actuated by independent tendons and each tendon can exert an arbitrary tension. These features can be favorable or undesired depending on the application, but the MTS design does provide a possibility of position coupling \textit{without} torque coupling.

\subsubsection{SA+MTS} This scheme is also heavily underexplored but of particular interest. Compared to TA+MTS which does not have position differential, the SA design in this scheme is advantageous since can provide \textit{position differential via tendon slack}: as the antagonistic tendon releases to enable joint motion by the spring, whenever the joint is stopped by the object, the tendon will get slack, which breaks the transmission from the motor and allows the motor (and thus other joints) to continue. This feature enables adaptive finger motion similar to the MJT paradigm. In addition, this design also has the precise position coupling, and also allow the decoupled joint torques, provided by MTS paradigm as explained above.
\section{Kinematic Design and Optimization}
\label{sec:conceptual_design}

We now show how the concepts introduced in the design matrix and discussed above can be used in concrete design problems. We assume the following design goals:
\begin{enumerate}
    \item A versatile hand must be able to execute a given number of grasps using as few motors as possible. This implies that postural synergies are used in order to shape the hand for each grasp, while torque synergies must ensure stability for each grasp.
    \item Once a grasp is executed, it must be maintained even if the motor is subsequently unpowered to save energy. 
    \item The hand must be backdrivable, i.e. allow itself to be closed or opened by external human intervention if no motor power is available.
\end{enumerate}
The specific application that drives these design goals is the development of a hand for an Assistive Free Flyer robot, Astrobee \cite{bualat2018astrobee}, on board the International Space Station. Such a hand must be versatile, able to interact with a range of objects for useful tasks; we thus selected a design with three fingers and eight joints. However, in order to fit inside the robot's payload bay, it must be highly compact, requiring the use of a single motor. Since battery power is limited, grasps must be maintained passively. Finally, for human-robot interaction purposes, the hand must be able to be closed or opened by external human intervention when powered off. Nevertheless, we believe that the capabilities we achieve here have broader applicability for different design cases.
    
In our previous work~\cite{chen2020underactuation} (briefly reviewed in Subsection \ref{subsec:opt}
), we introduced a design method to achieve goal (1) above by optimizing the tendon underactuation parameters, after (manually) selecting the tendon paradigm.

However, in our design examples in the previous work, like other designs, we exclusively used the TA+MJT tendon paradigm. Furthermore, we coupled abduction-adduction, proximal and distal joints in one finger along the same tendon, to facilitate the synergistic motion. However, this design forced the abduction-adduction torques and proximal and distal joint torques to be coupled, which is problematic: when the grasp is loaded, the proximal and distal joint torques should rise for grasping, but adduction torques are also undesirably increased. As a compromise (discovered by the optimization algorithm), we ended up with a tiny pulley for abduction-adduction joints, which, on the other hand, made the behavior sensitive to external disturbances (e.g. gravity, friction) and manufacturing errors. Furthermore, this paradigm does not offer any tools to achieve goals (2) and (3) from the list above. In this study, we extend our approach based on the design matrix introduced in the previous section to achieve these goals.

\subsection{Kinematic Design}

\begin{figure}[t]
    \centering
    \begin{subfigure}[t]{0.55\linewidth}
         \centering
         \includegraphics[width=\linewidth]{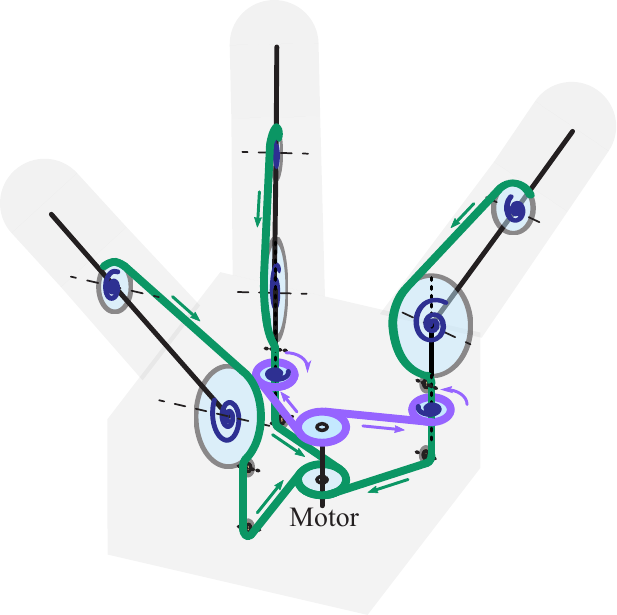}
         \caption{}
         \label{subfig:hand_tendon1}
    \end{subfigure}
    \hfill
    \begin{subfigure}[t]{0.35\linewidth}
         \centering
         \includegraphics[width=\linewidth]{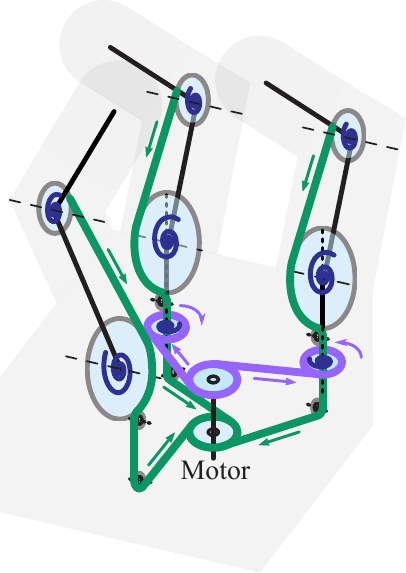}
         \caption{}
         \label{subfig:hand_tendon2}
    \end{subfigure}
    \caption{The tendon routing design of the proposed underactuated hand. (a) fully open, (b) partially closed. Green tendons are for finger flexion and purple tendons are for finger abduction-adduction. Please pay attention to the tendon winding directions and the abduction-adduction motion.}
    \label{fig:hand_conceptual_design}
    \vspace{-5mm}
\end{figure}

% The new design proposed in this paper is a successor of the aforementioned hands for the \textit{Astrobee} robot in ISS. However, the new design is aimed to perform versatile grasping instead of just perching on handrails. In addition, it is subject to the constraints of limited space, power and control bandwidth in ISS, and it should also take aforementioned energy and human safety concerns into consideration.

In the new design, we still use a three-finger eight-joint kinematic scheme (similar to \cite{chen2020underactuation}), where all fingers have proximal and distal joints for finger flexion and the fingers opposing the thumb also have abduction-adduction degrees-of-freedom (DoFs) (shown in Fig.\ref{fig:hand_conceptual_design}). We also actuate eight joints just by only one motor. 

% For the hand dimensions, we select them based on the sizes of objects of interest in ISS, and the size of the payload bay of \textit{Astrobee} robot in which the hand must be stowed. 

Our \textit{key innovations} in this design for the tendon transmission are as follows (illustrated in Fig.\ref{fig:hand_conceptual_design}) :

\begin{compactitem}
    \item For all finger proximal and distal joints, we choose the TA+MJT paradigm, in which one tendon goes through an idler in the proximal joint and wrap around and fixed to the distal joint pulley. 
    \item For the two abduction-adduction joints, we choose the SA+MTS paradigm, in which each joint connects to the motor via an independent tendon, and these joints are actuated by springs agonists. 
    \item On the motor shaft, the tendons for finger flexion (proximal and distal joints) and the tendons for abduction-adduction are winded in opposite directions, and when one set of tendons are collected, the others are released.
\end{compactitem}

This design can result in the following favorable features regarding underactuation and grasping behavior:

\begin{compactitem}
    \item The \textit{position differential (breakaway)} for different joints in a finger is achieved thus fingers can adapt to object shape mechanically --- both the TA+MJT in flexion (proximal and distal) joints and SA+MTS in abduction-adduction joints have position differential. Here we elaborate more on the breakaway behavior by the SA+MTS design in abduction-adduction joints: whenever adduction is blocked by objects, the tendons connected to these joints will get slack, while the proximal and distal joints can continue to flex the fingers.
    \item The \textit{position coupling} of all joints during free motion is achieved thus all joints can follow a predesigned trajectory driven by one motor. We note that, for SA+MJT joints, the position coupling is soft, while for SA+MTS this coupling is strict. Moreover, this design takes advantage of MTS paradigm to couple the abduction-adduction joints on different kinematic chains, illustrated in the second row of Table \ref{tab:design_matrix}.
    \item Finger \textit{abduction-adduction torques are not coupled with the flexion joints}, realized by the MTS design in abduction-adduction DoFs. The reason why this is essential is explained right before this subsection. 
    \item The \textit{grasping force regulation} via motor is achieved using the TA+MJT design across the flexion joints.
\end{compactitem}

\subsection{Optimization for Underactuation}
\label{subsec:opt}

The purpose of this optimization is to select the design parameters of the tendon underactuation mechanism. We take the methodology introduced in our previous work \cite{chen2020underactuation}. Here we provide a quick review of this work. Any underactuation mechanism corresponds to low-dimensional manifolds in joint posture and torque spaces. Our method first collects a set of sample grasps that span the useful joint space for grasping \textit{assuming independent joints (no joint coupling)}, then optimizes the underactuation parameters (pulley radii, spring stiffnesses, spring preloads) to shape these low-dimensional manifolds to span the sample grasps as closely as possible. For more details, we recommend the readers referring to \cite{chen2020underactuation} and \cite{chen2021interplay}.

We notice that the springs in abduction-adduction joints (SA+MTS) are not determined in the optimization. More generally speaking, unlike MJT, the springs on MTS joints do not affect the kinematic behavior of a finger. This gives us additional design freedom and opens a door to the benefits of spring force cancellation introduced in the next section.
\section{Spring Force Cancellation}
\label{sec:spring_cancellation}

\begin{figure}[t]
    \centering
    \includegraphics[width=0.7\linewidth]{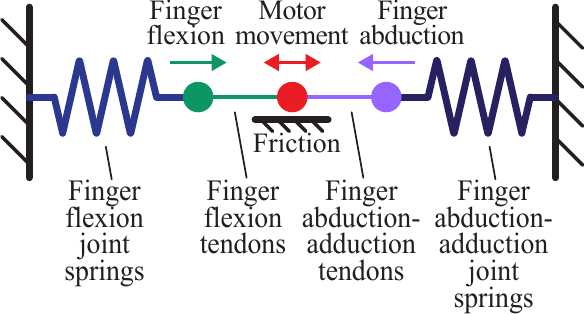}
    \caption{One-dimensional analogy of the tendon-spring system to illustrate the spring force cancellation. }
    \label{fig:spring_cancellation_1d_illustration}
    % \vspace{-5mm}
\end{figure}

\begin{figure}[t]
    \centering
    \includegraphics[width=\linewidth]{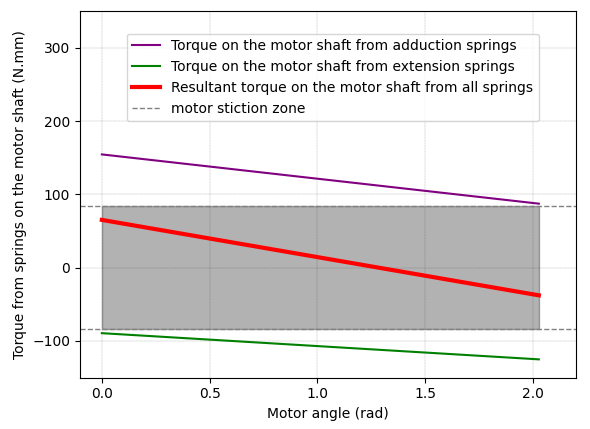}
    \caption{The plot of torques from springs on the motor shaft over the entire motor travel angles (motor angle 0: hand fully open; motor angle $2\pi /3 $ : hand fully closed).}
    \label{fig:resultant_motor_trq_plot}
    \vspace{-3mm}
\end{figure}

As discussed in Sec. \ref{sec:design_matrix}, the conventional TA+MJT design has great benefits, however, an unavoidable downside is that the motor always needs to fight against spring forces during grasping. By combining TA+MJT and SA+MTS design paradigms in one hand, we bring in the spring force cancellation (meaning the springs are  counterbalacing each other), which is advantageous in multiple aspects:

\begin{compactitem}
    \item Required motor torque can be reduced.
    \item Powering off the actuator subsequently will no longer lead to a release of the grasp. Instead, grasp is maintained in an energy-saving manner.
    \item With a reasonable external force from human intervention, the hand can be backdriven to create or release a grasp without motor power.
     
\end{compactitem}

These benefits make perfect senses in the ISS application discussed in Sec. \ref{sec:conceptual_design}. On one hand, the design can lower the hand energy consumption by passively maintaining grip even when powered off, which is especially useful for cases like perching on a handrail; on the other hand, it also allows astronauts to manually shape the hand to grasp or to release after powered off (using a hardware or software switch at will, or in case of power failure), as spring forces are mostly canceled out and a human only need to provide a force enough to break the gearbox stiction where no excessive effort is required.

In our design, we leverage agonist springs and antagonist springs on different joints to allow them to cancel each other by winding the flexion-extension tendons (green in Fig.\ref{fig:hand_conceptual_design}) and abduction-adduction tendons (purple in Fig.\ref{fig:hand_conceptual_design}) in the opposite directions on the motor shaft. To better demonstrate this idea, we illustrate the tendon-spring system in a simplified one-dimensional analogy in Fig.\ref{fig:spring_cancellation_1d_illustration}, showing the finger flexion and adduction springs counterbalancing each other via tendons.

As explained in Sec. \ref{sec:conceptual_design}.B, in the proposed design, the adduction springs are independent from the pre-contact kinematic behavior and thus not determined by the underactuation optimization algorithm, which provides extra design freedom. By carefully selecting the stiffnesses and preload angles of the abduction-adduction springs, we are able to reduce the resultant torque on the motor shaft to be lower than the maximum gearbox stiction and achieve the spring force cancellation over the entire range of motion, and the hand can hold its position even when powered off. 

Without contact forces, the resultant torque by springs on the motor shaft $\tau_{mot}^{net}$ can be calculated as: 
\begin{equation} \label{eq:net_torque}
\begin{split}
& \tau_{mot}^{net}  = \sum_i \frac{k_i  (\theta_i +\theta_{i}^{prel}) r_{mot}}{r_i} -  \sum_j \frac{k_j  (\theta_j +\theta_{j}^{prel}) r_{mot}}{r_j} \\
& i \in \{TP, F_1P, F_2P\}, j \in \{ F_1A, F_2A \}
\end{split}
\end{equation}
where $r$,$k$, $\theta$, $\theta^{prel}$ are joint pulley radius, joint spring stiffness, joint angle and joint spring preload angle respectively, and $r_{mot}$ is the motor pulley radius. $T, F_1, F_2$ refer to thumb, finger 1 and 2, and $P, A$ refer to proximal and adduction joint respectively. We note that the proximal and distal joints on the same finger share the same tendon force so the tendon force should be counted only once where we use proximal joint in the calculation. Also note that torque along the motor direction which leads to finger flexion is defined as positive and torque along the other way is thus negative.

Based on Equation (\ref{eq:net_torque}) and the availability of springs from the manufacturers, we select the adduction springs by their stiffness and design their preload angles to ensure that $|\tau_{mot}^{net}| < \tau_{mot}^{stiction}$ over the entire range of motion. In Fig. \ref{fig:resultant_motor_trq_plot}, we plot the torque from flexion-extension tendons (green), the torque from abduction/adduction tendons (purple), and the resultant torque (red) (all converted to the motor shaft). Considering the motor gearbox stiction in both motor driving directions, we mark the ``qualified zone'' with its upper and lower bound values being maximum gearbox stiction, shown as the gray highlight area in Fig.\ref{fig:resultant_motor_trq_plot}. Note that the red line is well enclosed by the "qualified zone" over the entire motor travel range, through which  spring force cancellation is guaranteed.

\section{Prototyping and Validation}
\label{sec:prototype}

\begin{figure}[t]
    \centering
    
    \begin{tabular}{ccccc}
         \includegraphics[width=0.22\linewidth]{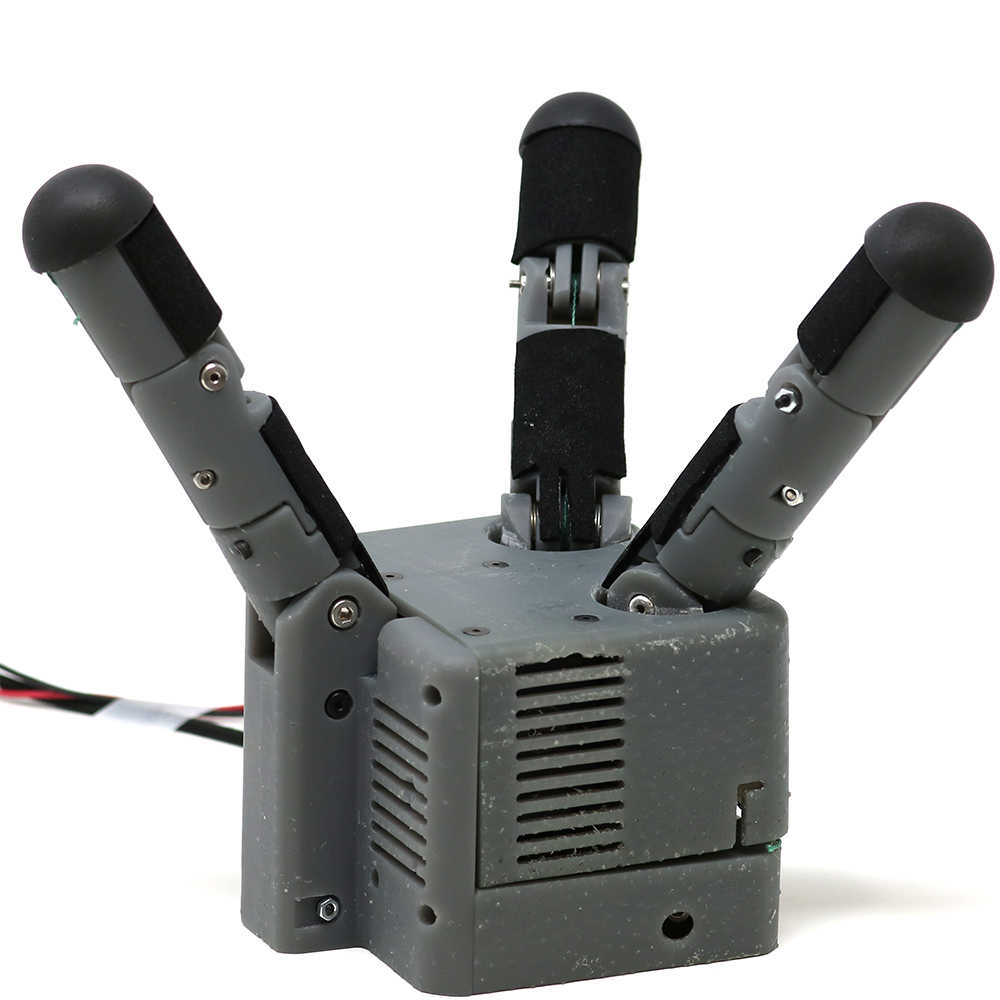}
         & \hspace{-5mm} \includegraphics[width=0.22\linewidth]{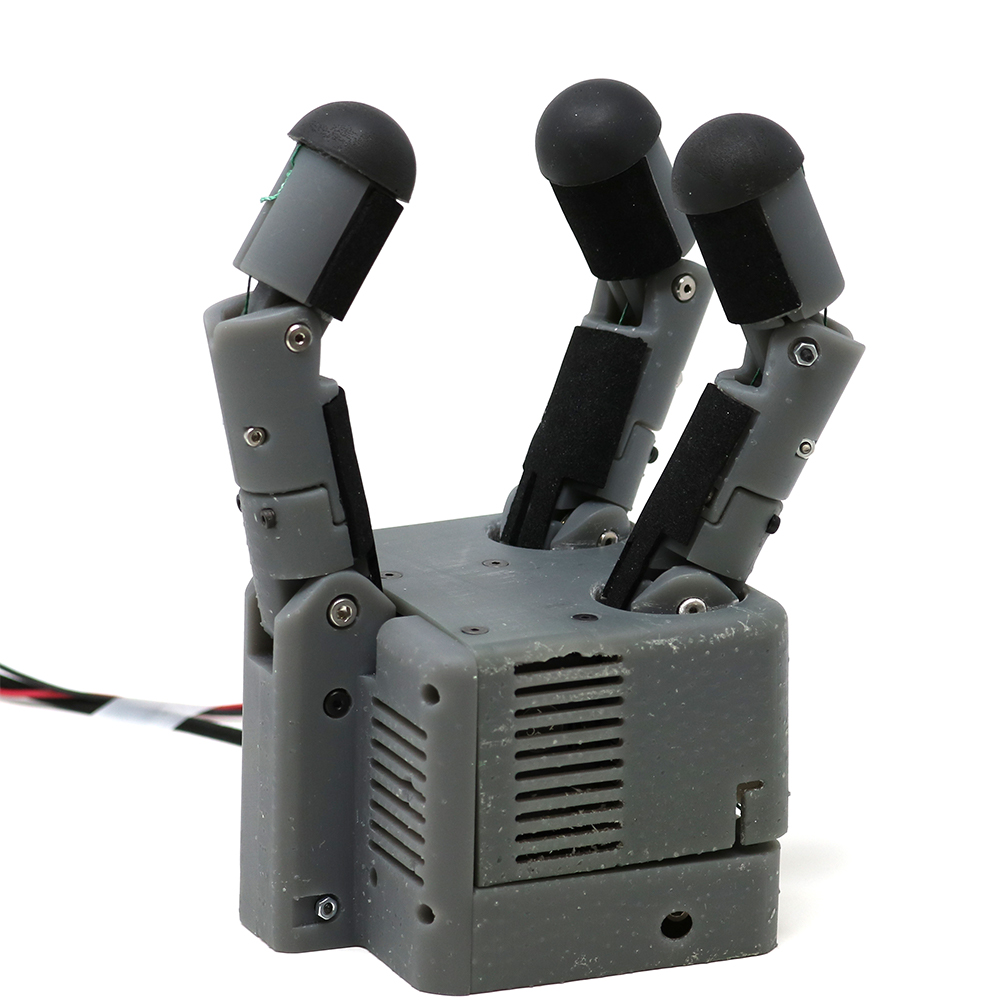}
         & \hspace{-8mm} \includegraphics[width=0.22\linewidth]{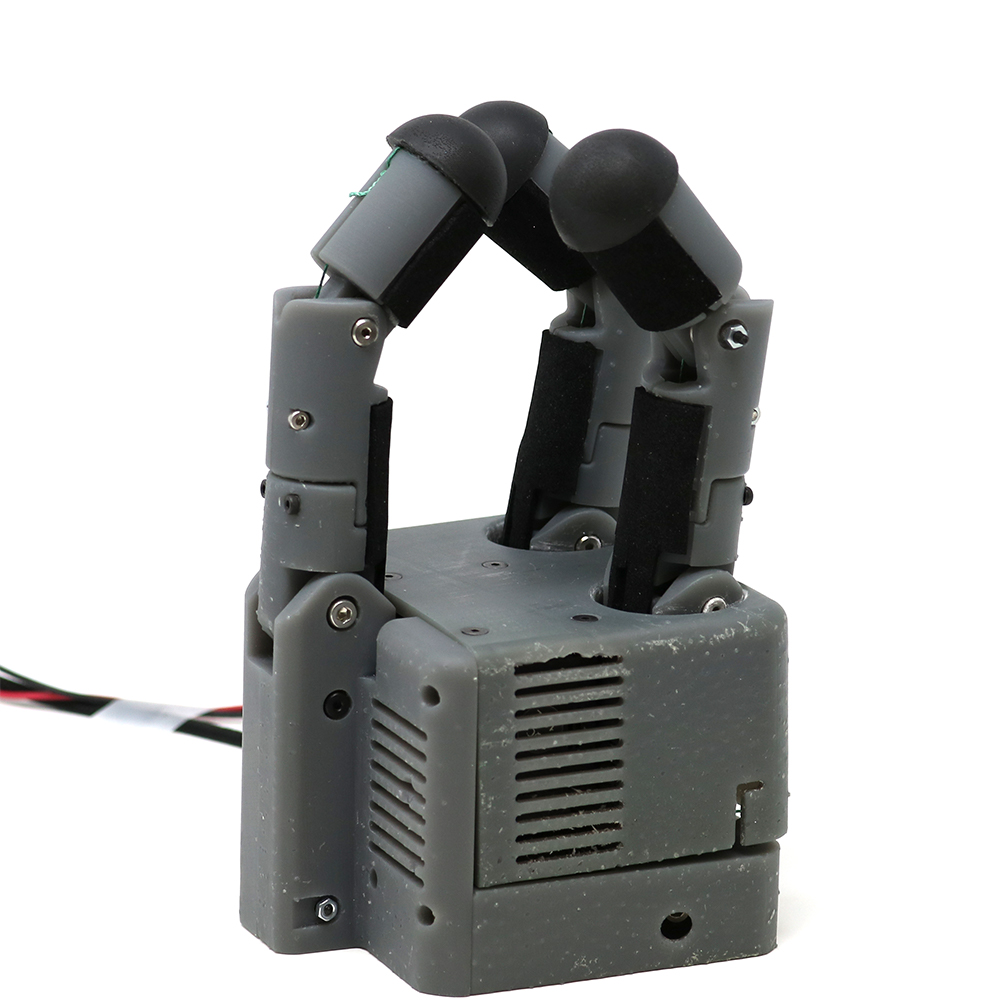} 
         & \hspace{-8mm} \includegraphics[width=0.22\linewidth]{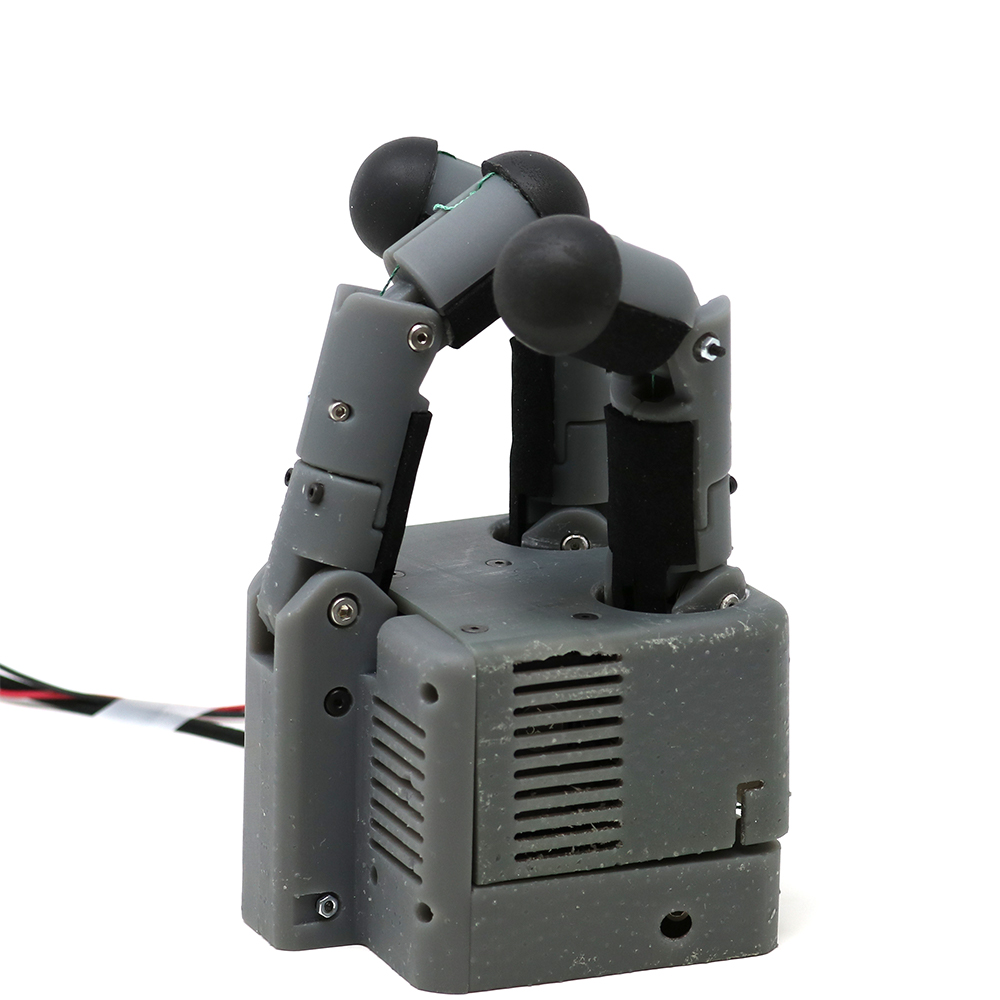} 
         & \hspace{-8mm} \includegraphics[width=0.22\linewidth]{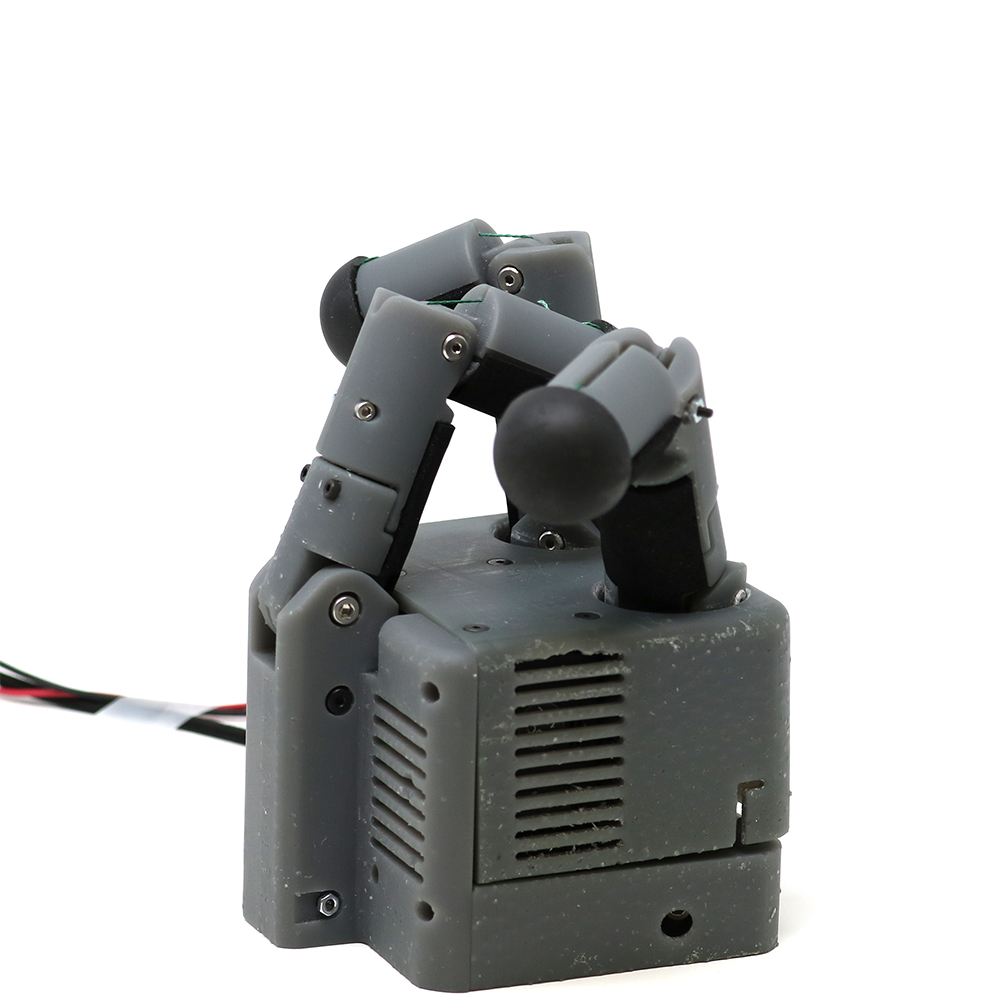}
    \end{tabular}
    \caption{The synergistic finger trajectories driven by the single actuator (from open to close).}
    \label{fig:hand_traj}
\end{figure}

\begin{figure}[t!]
    \centering
    \begin{subfigure}[b]{0.32\linewidth}
         \centering
         \includegraphics[width=\linewidth]{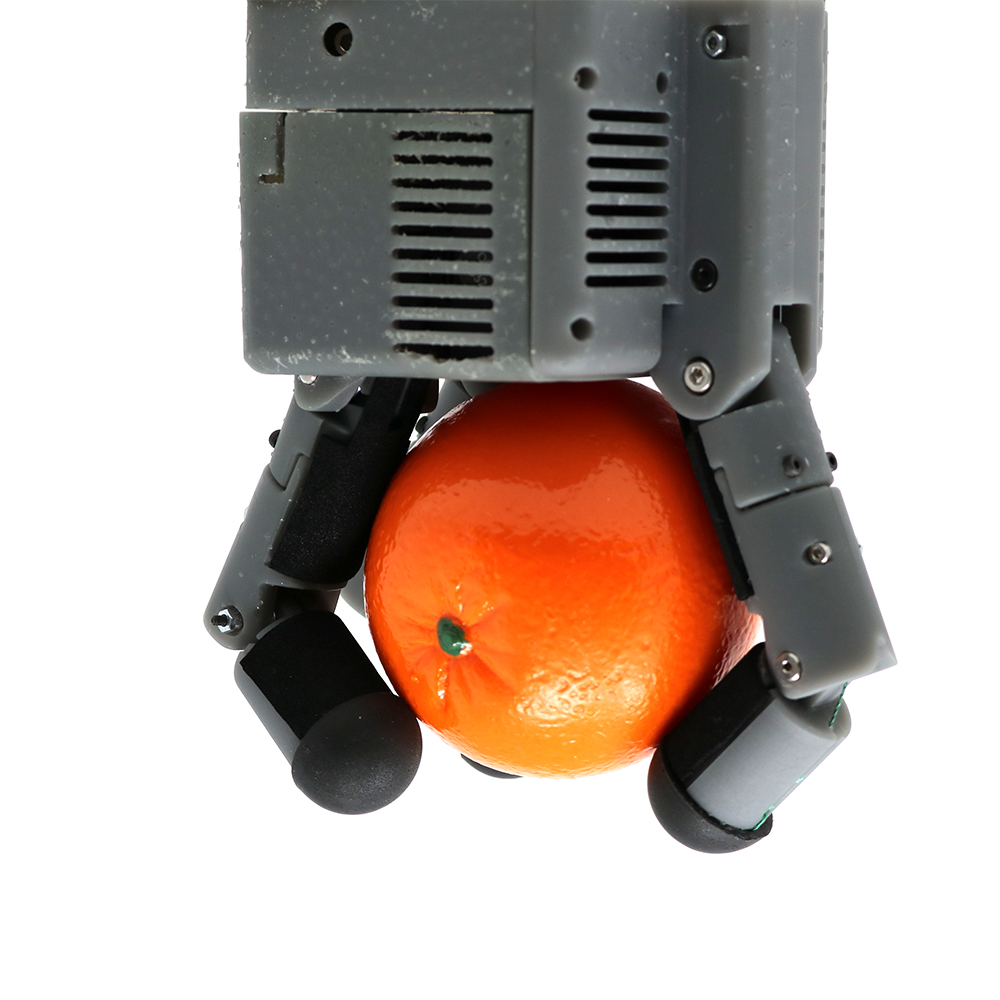}
        %  \caption{}
         \label{subfig:grasps_photo1}
    \end{subfigure}
    \begin{subfigure}[b]{0.32\linewidth}
         \centering
         \includegraphics[width=\linewidth]{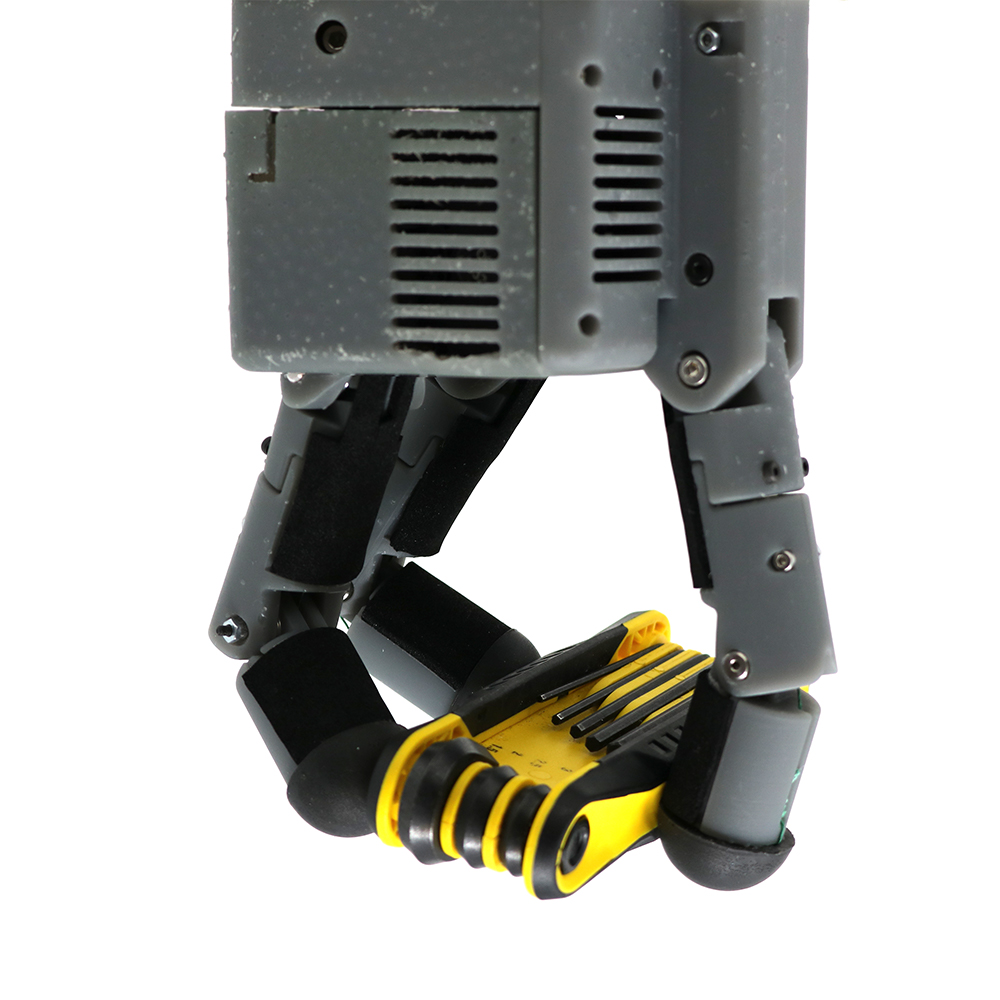}
        %  \caption{}
         \label{subfig:grasps_photo2}
    \end{subfigure}
    \begin{subfigure}[b]{0.32\linewidth}
         \centering
         \includegraphics[width=\linewidth]{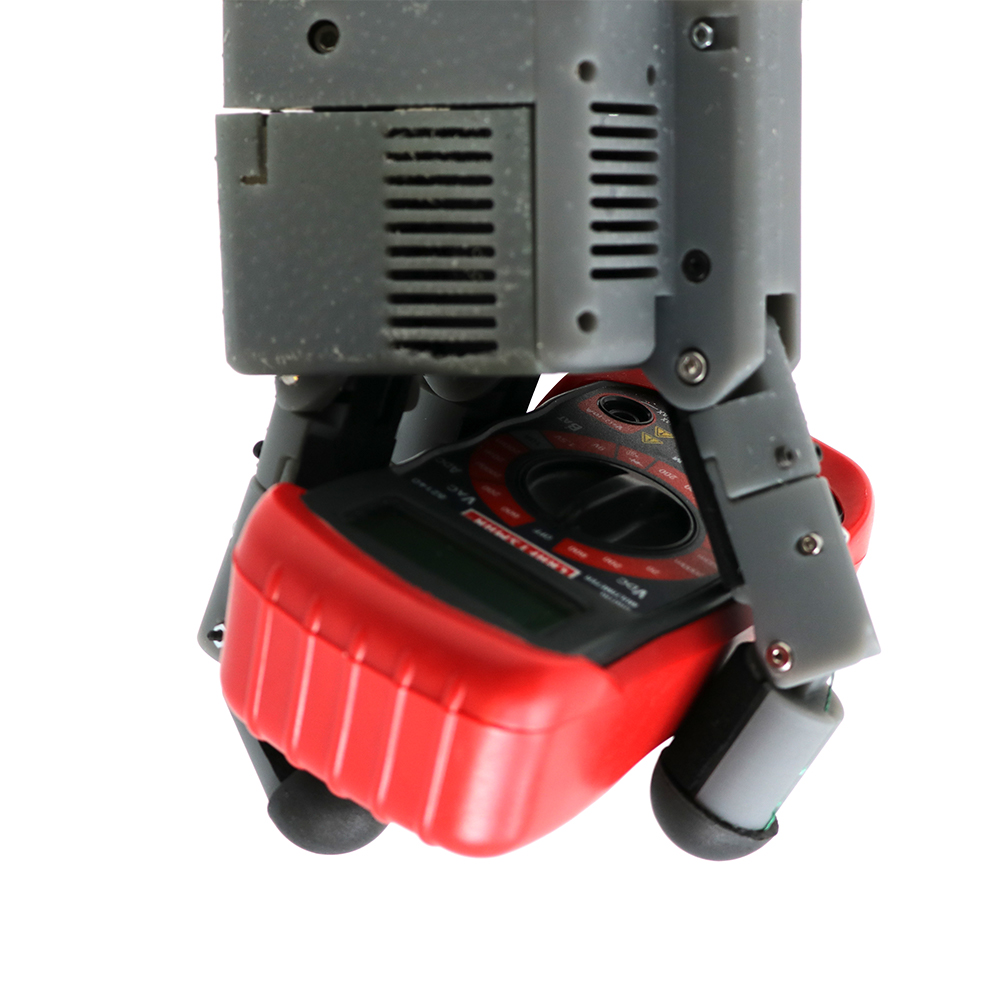}
        %  \caption{}
         \label{subfig:grasps_photo3}
     \end{subfigure}
     
     \vspace{-3mm}
    \begin{subfigure}[b]{0.32\linewidth}
         \centering
         \includegraphics[width=\linewidth]{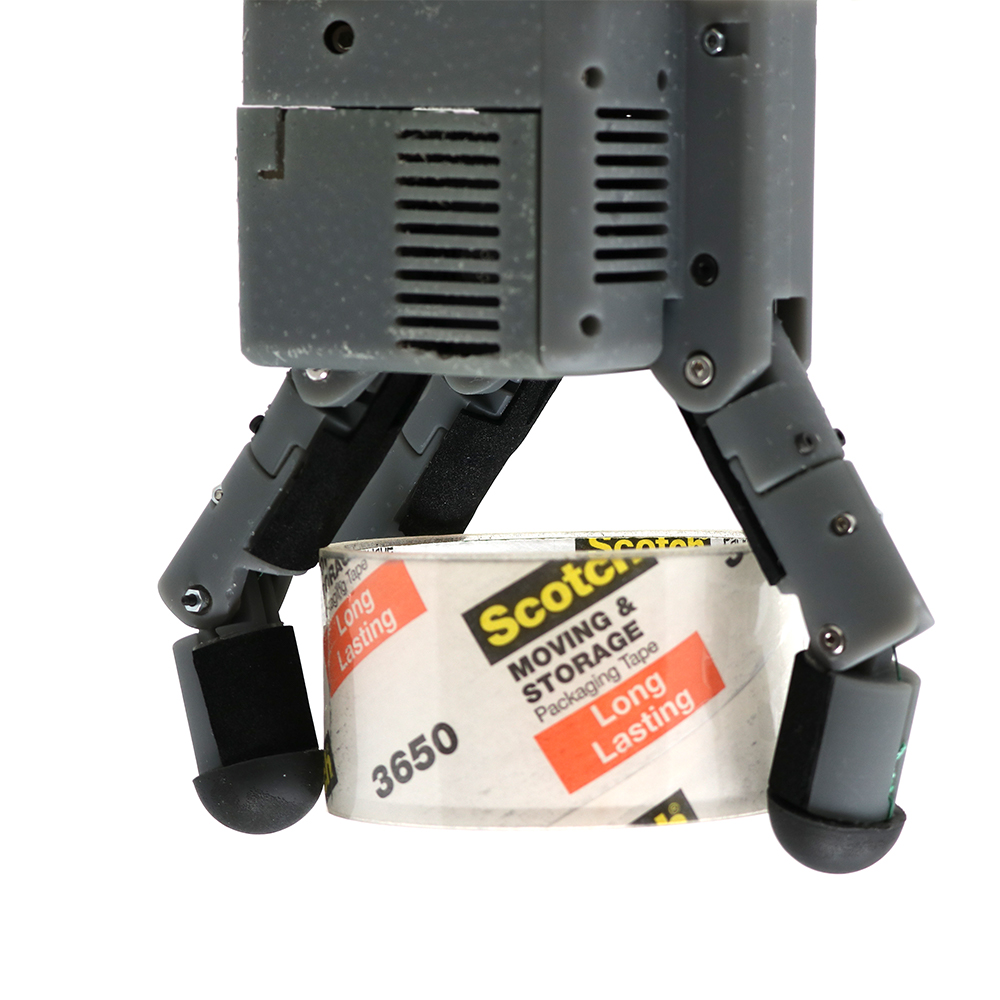}
        %  \caption{}
         \label{subfig:grasps_photo4}
     \end{subfigure}
    \begin{subfigure}[b]{0.32\linewidth}
         \centering
         \includegraphics[width=\linewidth]{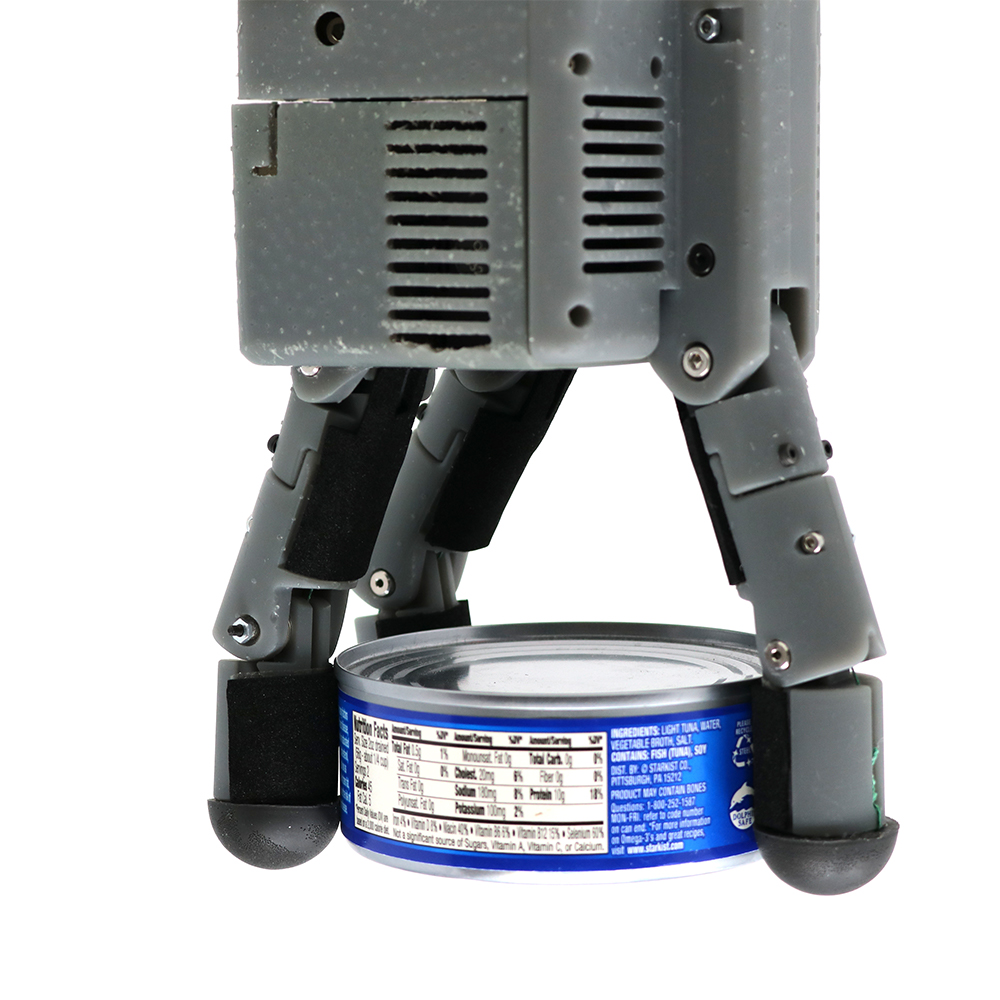}
        %  \caption{}
         \label{subfig:grasps_photo5}
     \end{subfigure}
    \begin{subfigure}[b]{0.32\linewidth}
         \centering
         \includegraphics[width=\linewidth]{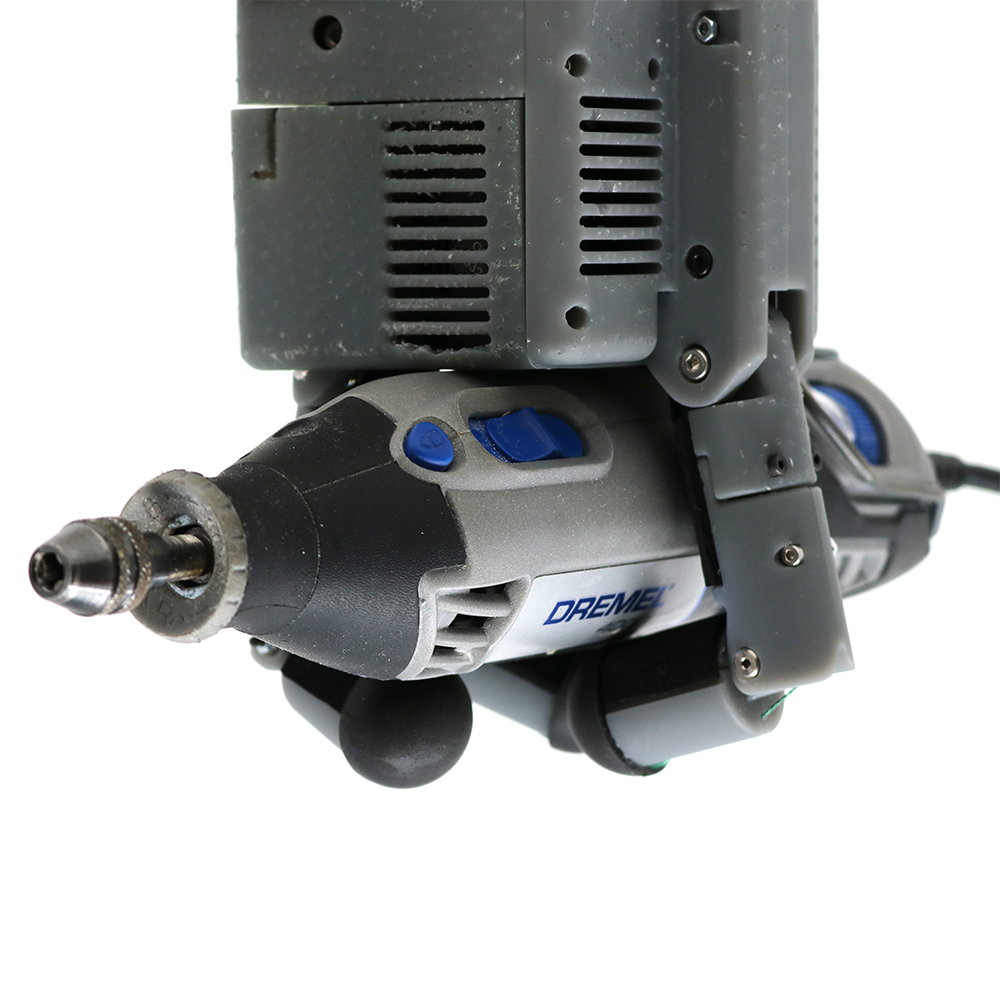}
        %  \caption{}
         \label{subfig:grasps_photo6}
     \end{subfigure}
     \vspace{-5mm}
    \caption{The example grasps with different objects.}
    \label{fig:example_grasps}
    % \vspace{-5mm}
\end{figure}

\begin{figure}[t]
    \centering
    \begin{subfigure}[b]{0.49\linewidth}
         \centering
         \includegraphics[width=0.8\linewidth]{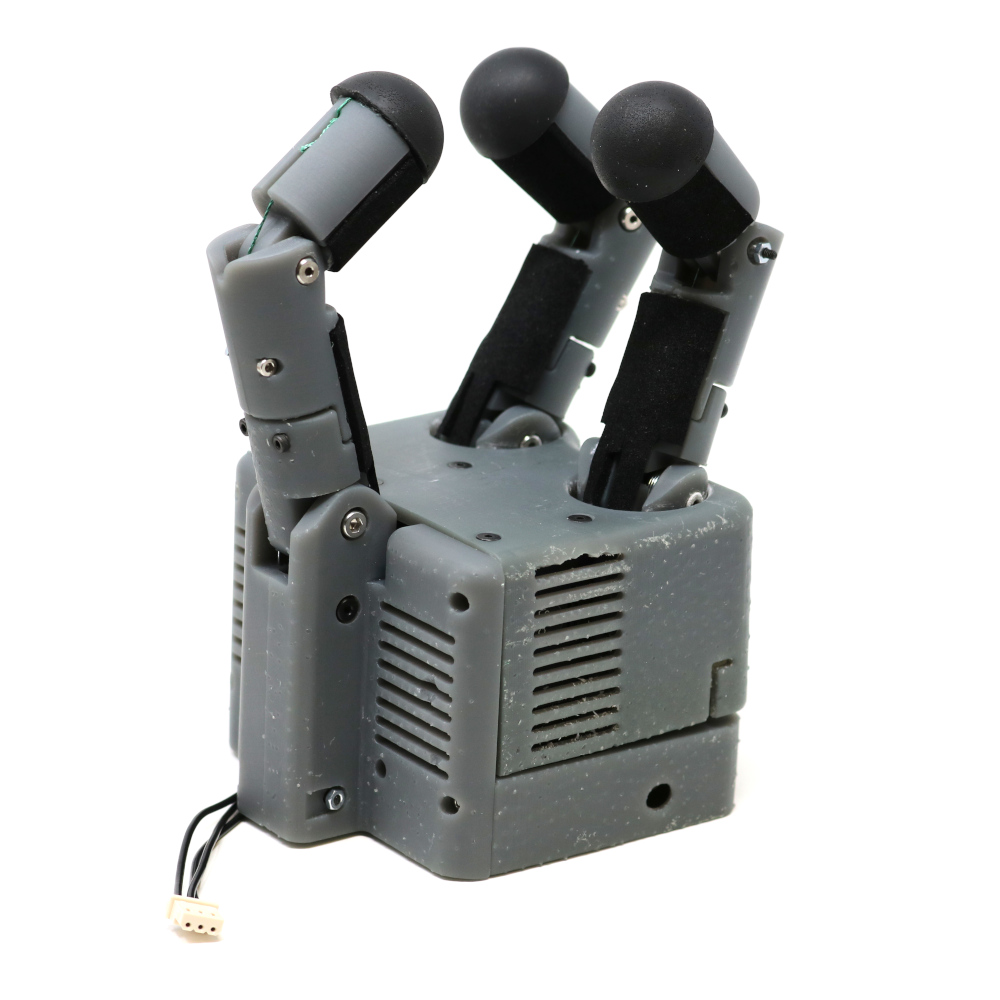}
         \vspace{-5mm}
         \caption{}
         \label{subfig:pose_keeping1}
    \end{subfigure}
    \begin{subfigure}[b]{0.49\linewidth}
         \centering
         \includegraphics[width=0.9\linewidth]{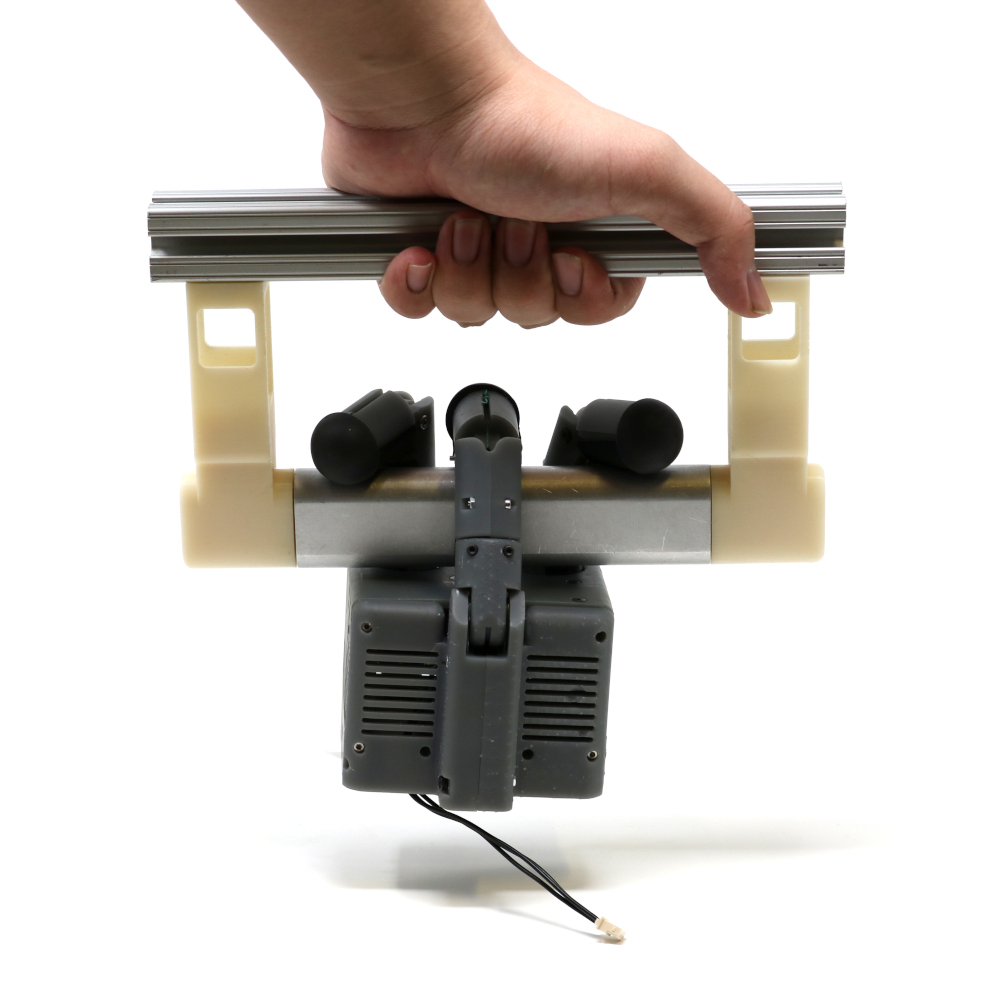}
          \vspace{-5mm}
         \caption{}
         \label{subfig:pose_keeping2}
    \end{subfigure}
     \caption{The power-off grip maintaining behavior in different hand configurations (note the power cable is unplugged). Particularly, (b) shows power-off perching on an ISS handrail.}
    \label{fig:pose_keeping}
\end{figure}

\begin{figure}[t!]
    \centering
    \begin{subfigure}[b]{0.49\linewidth}
         \centering
         \includegraphics[width=\linewidth]{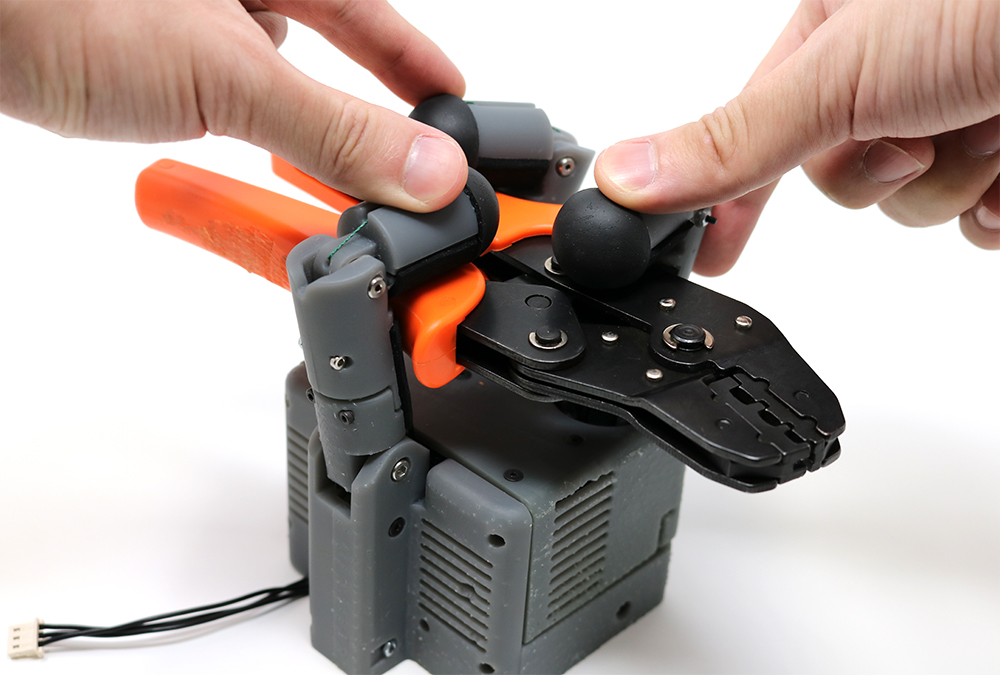}
         \caption{}
         \label{subfig:human_intervention1}
    \end{subfigure}
    \begin{subfigure}[b]{0.49\linewidth}
         \centering
         \includegraphics[width=\linewidth]{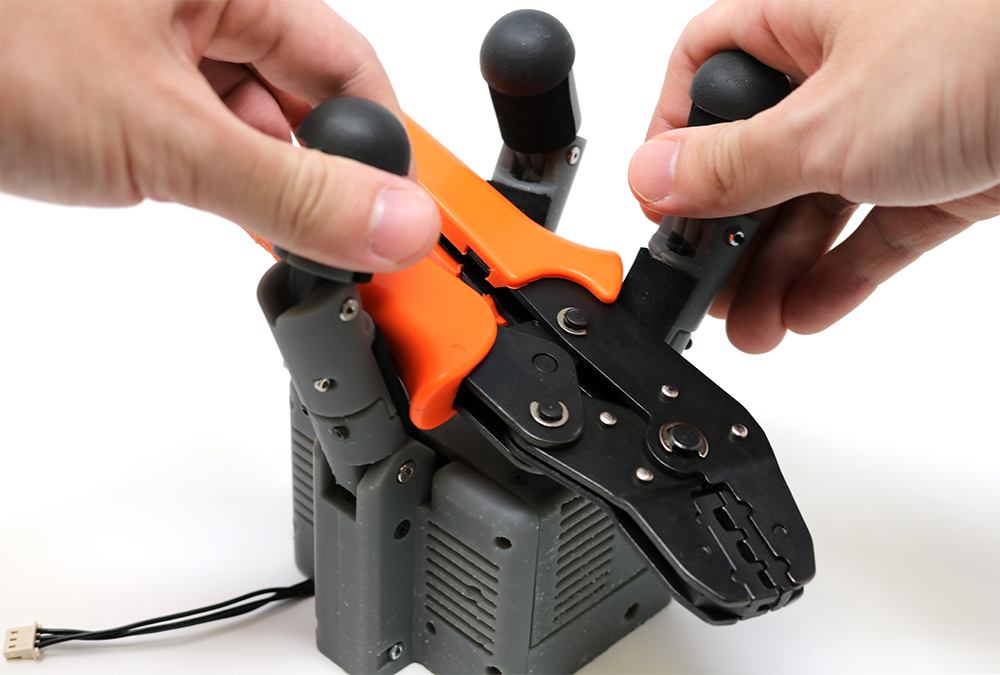}
         \caption{}
         \label{subfig:human_intervention3}
    \end{subfigure}
     \caption{Human intervention (when powered off): (a) manually close to grasp, (b) manually open.}
    \label{fig:human_intervention}
\end{figure}

\begin{figure}[t]
    \centering
    \includegraphics[width=\linewidth]{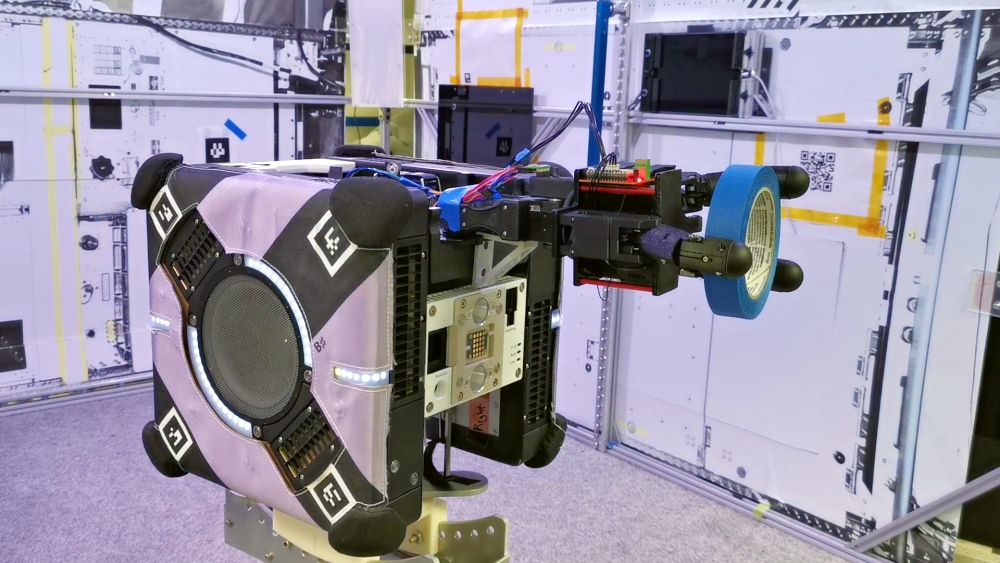}
    \caption{Preliminary integration tests with the \textit{Astrobee} free-flying robot.}
    \label{fig:Astrobee}
    \vspace{-5mm}
\end{figure}

We construct a physical prototype with the proposed design, shown in Fig. \ref{fig:hand_traj} and Fig. \ref{fig:example_grasps}. This hand is 3D-printed with resin (Formlabs Grey Pro), actuated by a single position-driven servo motor (Dynamixel XL430-W250-T), and controlled by an on-board microcontroller (PJRC Teensy 4.0) with custom circuits for controller-motor and controller-master communication.

Fig. \ref{fig:hand_traj} demonstrates the synergistic finger trajectories when the single motor is actuated. The hand closing start with a fully open position where the fingers are extended and abducted. Then the fingers can make spherical grasps for large ball-shaped objects. Then the fingertip can get close to create fingertip grasps for small objects. After that, since the abduction-adduction joints are moving with the flexion joints, the fingertips will not collide but will pass each other. Finally, the hand can make envelope grasps (power grasps) for long objects. Some example grasps are shown in Fig. \ref{fig:example_grasps}.

Experiment data shows that the gearbox in the servo motor can resist maximum 84 Nmm torque with its stiction. We selected appropriate abduction-adduction springs based on this value. As shown in Fig. \ref{fig:pose_keeping} and Fig. \ref{fig:human_intervention}, the hand can effectively keep pose after power-off, by virtue of the spring force cancellation. Especially, we show the unpowered perching on a ISS handrail in Fig. \ref{fig:pose_keeping} .

Shown in Fig. \ref{fig:human_intervention}, due to all-pose equilibrium after power-off, the hand can be manually shaped by humans to either make or release a grasp, providing operation convenience and human safety in the ISS application.

In collaboration with NASA, we were able to perform a preliminary integration test on an Astrobee Robot, as shown in Fig. \ref{fig:Astrobee}. We successfully set up the mechanical and electrical connection as well as communication to the \textit{Astrobee} robot. The hand prototype showed decent performance in common object grasping.

\section{Discussion and Conclusion}
\label{sec:discussion_conclusion}

In this paper, we discussed a design matrix regarding the tendon routing topology for tendon-driven underactuated hands, and proposed a hand design combining two useful design paradigms in this matrix, specifically TA+MJT for finger flexion and SA+MTS for abduction-adduction, which brings in various benefits in terms of synergistic grasping, energy consumption and human-robot interaction. 

Our experiments validate that the joint synergies are working as expected (as shown in Fig.\ref{fig:example_grasps}), while spring force cancellation is taking effect allowing energy saving and manual backdrivability when powered off (shown in Fig. \ref{fig:human_intervention}). We also note that the hand can create a stable grasp on the ISS handrail when powered off (shown in Fig.\ref{fig:pose_keeping}), enabling the power-off perching behavior similar to \cite{park2017developing} while significantly improved the grasping versatility.

For future work, on the engineering side, we aim to further integrate this hand with the \textit{Astrobee} robot and perform tests to validate the perching and grasping behaviors in a simulated ISS environment, e.g. with (simulated) micro gravity, with realistic vibrations, etc. On the design methodology side, we are also interested to investigate force regulation in SA design by introducing compliance in series with the tendon. We believe the concepts and practices introduced in this paper can help stepping towards the goal of versatile underactuated grasping, and we aim to explore new possibilities along this path.

%                                   % the textheight of the last page by a suitable amount.
%                                   % This command does not take effect until the next page
%                                   % so it should come on the page before the last. Make
%                                   % sure that you do not shorten the textheight too much.



%\section*{ACKNOWLEDGMENT}

%%%%%%%%%%%%%%%%%%%%%%%%%%%%%%%%%%%%%%%%%%%%%%%%%%%%%%%%%%%%%%%%%%%%%%%%%%%%%%%%

% References are important to the reader; therefore, each citation must be complete and correct. If at all possible, references should be commonly available publications.

\clearpage
\bibliographystyle{bib/IEEEtran}  
\bibliography{bib/analysis,bib/biomechanics,bib/control,bib/design,bib/learning,bib/planning,bib/sensing,bib/simulation}

\end{document}